\documentclass[letterpaper]{article}
\usepackage{aaai}
\usepackage{times}
\usepackage{helvet}
\usepackage{courier}
\usepackage{hyperref}
\usepackage{graphicx}
\usepackage{comment}
\usepackage{subcaption}
\usepackage{stfloats}

\usepackage{algpseudocode}
\usepackage{xcolor}
\usepackage[linesnumbered,ruled,vlined]{algorithm2e}

\SetCommentSty{mycommfont}

\SetKwInput{KwInput}{Input}                
\SetKwInput{KwOutput}{Output}              

\begin{document}
%
\title{Say ``Sul Sul!\thanks{\emph{Sul sul} means \emph{hello} in Simlish, the fictional language spoken in The Sims.}'' to SimSim, A Sims-Inspired Platform for Sandbox Game AI}

\author{Megan Charity \\  Tandon School of Engineering \\ New York University  \\ mlc761@nyu.edu \And
Dipika Rajesh \\ Tandon School of Engineering \\ New York University \\ dr2898@nyu.edu \And
Rachel Ombok \\ Tandon School of Engineering \\ New York University \\ rbo224@nyu.edu \And
L. B. Soros \\ Tandon School of Engineering \\ New York University \\ lsoros@nyu.edu}

\maketitle
\begin{abstract}
\begin{quote}
This paper proposes environment design in the life simulation game The Sims as a novel platform and challenge for testing divergent search algorithms. In this domain, which includes a minimal viability criterion, the goal is to furnish a house with objects that satisfy the physical needs of a simulated agent. Importantly, the large number of objects available to the player (whether human or automated) affords a wide variety of solutions to the underlying design problem. Empirical studies in a novel open source simulator called SimSim investigate the ability of novelty-based evolutionary algorithms to effectively generate viable environment designs. 
\end{quote}
\end{abstract}

\section{Introduction}




Sandbox games require constrained creativity and exploring complex design spaces while avoiding nonviable portions. Examples of such games include Minecraft\footnote{Copyright (c) 2011 Mojang}, in which players must build shelters out of natural resources, and SimCity\footnote{Copyright (c) 1989 Electronic Arts, Inc}, which requires players to set zoning policies and otherwise allocate resources to enable simulated city population growth. This genre is unique because success can be achieved using an enormous variety of strategies. While this relative freedom creates a unique gaming experience for human players, it also presents an underexplored challenge for game AI: finding a diversity of creative artifacts that satisfy a set of minimal constraints. Such domains can serve as an apt testbed for an underexplored class of divergent evolutionary search algorithms that are driven primarily by binary minimal fitness criteria instead of continuous fitness gradients.

\begin{figure}[!h]
\centering
  \includegraphics[width=0.45\textwidth]{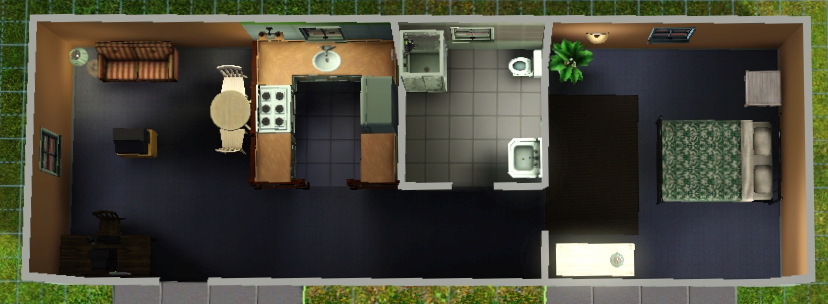}
  \caption{\textbf{A human-designed house in The Sims 3.} The SimSim simulator simplifies the design problem into that of designing a single functional room (with each furniture object occuping a single tile), though the simulator also supports evolving entire houses.}
  \label{fig:house}
\end{figure}

In The Sims ((c) 2008 Electronic Arts, Inc.), the player controls simulated humans called Sims that inhabit a house, which can be either preconstructed or built and furnished by the player. An example human-built house is shown in Figure \ref{fig:house}. Each Sim has a set of needs (Hunger, Energy, Social, Fun, Bladder, and Hygiene) that must be satisfied, with satisfaction levels decreasing over time. Interactions with objects, such as furniture, and with other Sims can have positive and negative effects on need satisfaction levels. The primary player task is then to select actions that will prevent the Sim's needs from becoming critically low. Importantly, there are no ``win'' conditions beyond any personal goals that the player might set. However, a Sim can die if certain needs go unmet for too long.
Thus, The Sims contains two separate yet related gameplay challenges: 1) designing a house with sufficient objects to prevent a Sim's needs from becoming critically low, and 2) selecting interactions with these objects such that the Sim's decaying need satisfaction levels are sufficiently replenished.

This paper introduces a new open source simulator based on house design in The Sims and demonstrates its utility for exploring minimal-criterion-based search algorithms. Though prior work by e.g. \citeauthor{yu2011make} (\citeyear{yu2011make}) has explored optimizing furniture arrangement for realistic living spaces, the goal herein is to discover quality diversity in the context of minimal viability criteria. The new experimental domain and accompanying source code are the primary contributions of this paper, in addition to arguing more generally that sandbox games are interesting testbeds for AI and, more specifically, for divergent search algorithms.
First, existing work on this class of algorithms is reviewed. The SimSim Sims simulator is then introduced and key experiment-enabling differences from the original game are noted. Experiments with variants of an algorithm called minimal criterion novelty search are then reported, highlighting the novel platform's utility for future studies on divergent evolutionary search algorithms and procedural content generation.

\section{Background}


Evolutionary computation (EC) is a class of biologically-inspired algorithms encoding points in a search space as genomes (i.e. mutatable parameter vectors). 
While EC methods differ in terms of genome representation (and potentially other features), they all share a common approach of exploring a search space by mutating and sometimes combining genomes representing already-discovered individuals. The general idea, motivated by the concept of ``survival of the fittest'', is to iteratively create a population of genomes, score them on some fitness-bearing domain, and then select the elites as parents for the next generation. 

\citeauthor{lehman:ecj11} (\citeyear{lehman:ecj11}) turned this idea on its head, demonstrating that novelty alone, without fitness-based selection, could solve a maze navigation problem not solvable by fitness-based search alone. This idea was later extended by minimal criteria novelty search (MCNS) \cite{lehman:gecco10a}, which increases search efficiency by requiring that individuals satisfy some minimal fitness threshold in order to reproduce.
This algorithm was transformative because it emphasized ``survival of the fittest'' instead of ``survival of the fit \emph{enough}'', allowing more room for evolutionary creativity. Further innovations on MCNS include progressive minimal criterion novelty search \cite{gomes2012progressive}, evolution in Chromaria \cite{soros:alife14}, minimal criterion coevolution \cite{brant2017minimal}, and POET \cite{wang2019poet}. 

Novelty search, in part inspired a recently popular class of evolutionary algorithms called quality diversity (QD) algorithms that aim to discover a wide variety of high-performing genomes within a particular search space. Most QD algorithms derive from either novelty search with local competition (NSLC)\cite{lehman:gecco11} or MAP-Elites \cite{mouret:arxiv15}. 
QD algorithms have also been used for procedurally generated game content such as spaceships \cite{liapis2013transforming}, strategy game maps \cite{liapis2015constrained}, bullet hell game levels \cite{khalifa2018talakat}, and RPG dungeons \cite{alvarez2019empowering}. \citeauthor{gravina2019procedural} (\citeyear{gravina2019procedural}) give a review of QD+PCG work.

Though intentionally divergent, QD algorithms are frequently cast as optimizers. More critically, the domains used to evaluate them rarely capture the prolific creative potential of evolutionary generative systems. 
Initial experiments focused on single-objective maze navigation tasks. 
Subsequent maze domains involved more complicated mazes explicitly designed with multiple viable paths to the end point \cite{pugh:frontiers16}, shifting the goal to finding \emph{all} such paths, not just one. Artificial life domains for QD have largely focused on morphologies or gaits for virtual creatures or physical robots. Games such as Hearthstone \cite{fontaine2019mapping} have also been used as testbeds for QD algorithms. Still, most of these domains naturally implement a gradient-based fitness function, enabling ``survival of the fittest'' instead of ``survival of the fit enough''. 

The novel sandbox-game-based simulator described in the next section implements a natural minimal fitness criterion, creating an important new testbed well-suited to this important new class of algorithms.
A few prior Other works have posed sandbox games as challenges for AI in general. For instance, \citeauthor{earle:exag2020} (\citeyear{earle:exag2020}) explored reinforcement learning in SimCity. Minecraft has become an increasingly popular domain for creative AI research in games. In the Generative Design in Minecraft (GDMC) competition \cite{salge2020ai}, the primary challenge is for an algorithm to generate plausible settlements on arbitrary terrains. Submissions are scored by human judges with respect to adaptation, functionality, evocative narrative, and aesthetics. Importantly, while the theme of generative design is shared between GDMC and SimSim, and while GDMC implicitly rewards algorithms that can generate a diversity of high-quality designs (by evaluating submitted algorithms on multiple terrains), GDMC does not incorporate the survival aspect of Minecraft and thus is not an ideal testbed for minimal-criterion-based algorithms. Inversely, the MineRL competition\footnote{https://minerl.io/competition/} focuses on high-performing agent gameplay behaviors, with little emphasis on creativity. \citeauthor{soros:gecco17} (\citeyear{soros:gecco17}) introduced a simplified Minecraft domain called Voxelbuild explicitly designed for evaluating divergent evolutionary search algorithms. However, this system removes the Minecraft survival mechanic, thereby removing the minimal viability criterion from the game.

\section{SimSim: A Sims Simulator}



The SimSim simulation\footnote{Code: \url{https://github.com/lsoros/simsim}.} replicates select game mechanics from The Sims for the purpose of exploring algorithms that find a diversity of high-quality designs with minimal viability criteria. This section first describes game mechanics in The Sims before noting how SimSim is different.


In The Sims, a human player commands one or more Sim agents to navigate to particular objects and interact with them, replenishing one or more needs. Sims can also select their own goals probabilistically; when not controlled by the player, Sims use A* pathfinding to move towards one of the four highest-ranked objects according to impact on ``happiness'' (combined Sim needs ordered by Maslow’s Hierarchy of Needs). This decision-making process can also be influenced by distance to objects, personality motives, interaction with other Sims, and mood \cite{bourse_2012}. 

 In SimSim, a single agent (Algorithm \ref{alg:sim_agent}) lives in one-room houses and interacts with furniture objects. The simulator contains 70 hand-coded objects, each with a vector of effects on Sim needs. For the experiments in this paper, objects are placed by evolutionary algorithms, but other strategies could be implemented. There is no player interaction, and the Sim has full knowledge of every object in the room. Needs are prioritized based on a preset order, so that the Sim will attend to a more critical need (i.e. hunger or energy) over others. (SimSims die if Hunger or Energy reaches 0.) 
 Navigation uses breadth-first search in a room with X-Y positive integer coordinate values. The agent can take one step up, down, left, or right each tick. Once the agent reaches its target object, it can interact with it. SimSims interact with objects until the relevant diminished need is satisfied. Some objects replenish a need more than others per interaction, but it is unlikely for all needs to be maximally satisfied. SimSim only allows one agent at a time and social interactions are not implemented beyond some objects satisfying the Social need. SimSim agents also only have one possible generic interaction with each object, instead of selecting from options that might each affect the Sim’s needs differently. SimSim happiness is a simple combination of all needs, whereas happiness in The Sims is impacted more strongly by unfulfilled needs than fulfilled ones. 
 


\begin{algorithm}[!ht]
\DontPrintSemicolon
\KwInput{maxTicks, dimRate, needsRankings, thresh}
\KwOutput{fitness}
\While{$curTick < maxTicks$}{
    \tcc{Check if Sim agent is dead}
    \If{$hunger \leq 0$ or $energy \leq 0$}{break}
    \tcc{Set target path if not yet set}
    \If{($\exists targetObject$) and ($\neg \exists targetPath$)}{
        $targetPath$ = bfs($targetObject$)\;
    }
    
    \tcc{Move to target and interact}
    goto next path point\;
    \If{sim at $targetObject$ position}{
        Interact with $targetObject$\;
        $targetObject$ = null\;
    }
    
    \tcc{Diminish needs based on tick}
    \For{$i$ in (\# sim needs)}{
        \If{(dimRate[$i$] \% $curTick$) = 0}{
            simNeeds[$i$] -= 1\;
        }
    }
    
     \tcc{Set the next target object}
     \If{$\exists$ $targetObject$}{
         continue\;
     }
    \For{$need$ in $needsRanking$}{
        \If{simNeeds[$need$] $<$ thresh}{
            $targetObj$ = findNearestObj($need$)\;
        }
    }
    
     $curTick$++\;
}

return $fitness$\;

\caption{SimSim Agent}
\label{alg:sim_agent}
\end{algorithm}

\section{Methodology}
This paper explores the impacts of varying the fitness threshold for minimal criterion novelty search in a novel sandbox game domain that naturally implements a minimal fitness criterion (i.e. avoiding death), unlike most other domains for testing similar algorithms. More importantly, these experiments demonstrate the viability of the SimSim simulator for future work on quality diversity algorithms. For these experiments, evolution begins from a single seed level that meets some, but not all, Sim needs. Figure \ref{fig:minroom} depicts this basic room as rendered in The Sims 3. Note that SimSim objects actually only occupy one tile each. The challenge for evolution is to find objects that satisfy the Sim's needs and also don't kill the Sim (by i.e. removing objects necessary for basic survival without replacing them with objects satisfying equivalent needs). 
\\

\noindent The following algorithms are compared:

\begin{figure}
\centering
  \includegraphics[width=0.25\textwidth]{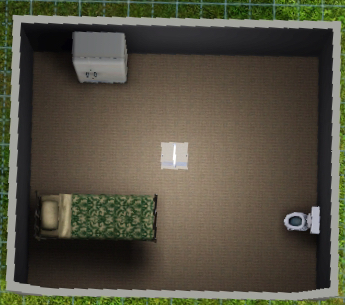}
  \caption{\textbf{Minimal initial seed room for evolution.} If the evolutionary algorithm removes any objects in this room without replacing them with objects satisfying equivalent needs, the Sim cannot survive, thereby earning 0 fitness.}
  \label{fig:minroom}
\end{figure}

\begin{itemize}
    \item \textbf{Novelty search (NS)}, shown in Algorithm \ref{alg:novel_search}. The objects in each room are encoded as vectors of objects. The vector distance between rooms in the population and novelty archive is then calculated and used for the k-nearest neighbor calculation in the vector space. A generated house is considered novel if the distance to previously discovered houses exceeds a fixed threshold. This new house is then added to the novelty archive. 
    
    \item \textbf{Minimal Criterion Novelty Search (MCNS)}. In this modification to novelty search, individuals in the population (i.e. rooms) are given a fitness of 0 if their earned fitness is $<$ 0.1. This algorithm is motivated by the binary conception of fitness observed in biological evolution.
    
    \item \textbf{Minimal Criterion Novelty Search with higher fitness threshold (MCNS-H)}. In this case, fitness is reduced to 0 if an individual earns $<$ 0.2 fitness.
    
    \item \textbf{(1+1) Evolutionary Algorithm}, shown in Algorithm \ref{alg:easearch}. This algorithm was selected as an example of a canonical greedy evolutionary algorithm, also known as a ``hillclimber'', that optimizes solutions to a problem using extremely incremental local search, and is a standard comparison algorithm for evolutionary computation research. The main purpose of including it in experiments herein is to see if the domain forces algorithms to sacrifice quality for the sake of diversity. The population consists of one parent and its single child, with the more fit of the two serving as the sole parent of the next two-member generation.
\end{itemize}

\begin{algorithm}[!ht]
\caption{Novelty Search}
\label{alg:novel_search}
 \KwInput{$popSize, generations$}
    $curGen$ = 0\;
    $initHouse$ = House(toilet, bed, fridge)\;
    $pop$ = []    \tcp*{house population}
    $novPop$ = [] \tcp*{novelty archive}
    \For{$i$ in $populationSize$}{
        $mutHouse$ = mutate($initHouse$)\;
        add $mutHouse$ to $pop$\;
    }
    \tcc{Start the evolution}
    \While{$curGen < generations$}{
        \tcc{Run game and check novelty}
        \For{$h$ in $pop$}{
            $testSim$ = SimAgent()\;
            $fitness$ = simulate($h$,$testSim$)\;
            \If{isNovel($h$,$fitness$,$novPop$)}{
                add $h$ to $novPop$\;
            }
        }
    
        \tcc{Randomly select houses from novel archive and current population, mutate the parents, and add to the new population}
        $newPop$ = []\;
        $s$ = $popSize$/6\;
        $novParents$ = random($novPop$, $s$)\;
        \For{$n$ in $novParent$}{
            \For{i in 3}{
                $newHouse$ = mutate($n$)\;
                add $newHouse$ to $newPop$\;
            }
        }
        $popParent$ = random($pop$, $s$)\;
        \For{$p$ in $popParent$}{
            \For{i in 3}{
                $newHouse$ = mutate($p$)\;
                add $newHouse$ to $newPop$\;
            }
        }
        
        \tcc{Replace old population}
        $pop$ = $newPop$\;
        $curGen$++\;
    }
    \tcc{Export the novelty archive houses}
    exportToJSON(novPop)\;
    
\end{algorithm}

\begin{algorithm}
\caption{(1+1) Evolutionary Search}
\label{alg:easearch}
\KwInput{$generations$}
    $curGen$ = 0\;
    \tcc{Create initial house and get the fitness}
    $initHouse$ = House(toilet, bed, fridge)\;
    $pop$ = []    \tcp*{house population}
    $startSim$ = SimAgent()\;
    $initFitness$ = simulate($h$,$startSim$)\;
    
    \tcc{Set the initial best house}
    $bestHouse$ = $initHouse$\;
    $bestFitness$ = $initFitness$\;
    
    \tcc{Start the evolution}
    \While{$curGen < generations$}{
        \tcc{Copy and mutate the best house}
        $childHouse$ = mutate(House($bestHouse$))\;
        $childSim$ = SimAgent()\;
        $childFit$ = simulate($childHouse$,$childSim$)\;

        \tcc{Replace the best house with the better child}
        \If{$childFit > bestFitness$}{
            $bestHouse$ = $childHouse$\;
            $bestFitness$ = $childFitness$\;
        }
        
        $curGen$++\;
    }
    \tcc{Export the best house}
    exportToJSON(bestHouse)\;

\end{algorithm}




\noindent Regardless of the algorithm being tested, the fitness value of a room is based on the total need values divided by the maximum total possible for all the values. The following formula represents the fitness value calculated:
\begin{equation}
    f = \frac{b+f+h+s+e+y}{60}
\end{equation}
where $f$ is the final fitness value and $b$, $f$, $h$, $s$, $e$, and $y$ are the final Bladder, Fun, Hunger, Social, Energy, and Hygiene values respectively. If a Sim ``dies'' before the end of the ticks, the fitness value for the house is set to 0. The maximum fitness value is 1.



The room genome is a list of furniture objects contained in the room. While not yet explored in this paper, multi-room houses can also be created, and they are encoded as lists of rooms. Room mutations include moving an existing object, deleting an existing object, and adding a new object, with probabilities 0.40, 0.10, and 0.30 respectively. New objects are added at random positions. If there is a collision with an existing object at the selected location, the system will attempt to place the new object in a neighboring location, but will abort mutation if all neighboring tiles are occupied.   

Each algorithm was evaluated over 20 runs, each through 100,000 generated individuals. Population size for all NS and MCNS runs was 100, with 1,000 generations. This end time was chosen based on a smaller set of preliminary experiments through 500,000 individuals, which showed that the algorithms tended to converge by 100,000 individuals. (1+1)-EA runs also lasted through 100,000 individuals, but population size was 2. The minimum K-nearest neighbor distance was 3. House size was $7x7$ tiles and Sims lived for 100 ticks.  


\section{Results}

\begin{figure*}[t]
    \centering
    \begin{subfigure}[t]{.33\linewidth}
        \includegraphics[width=0.95\linewidth]{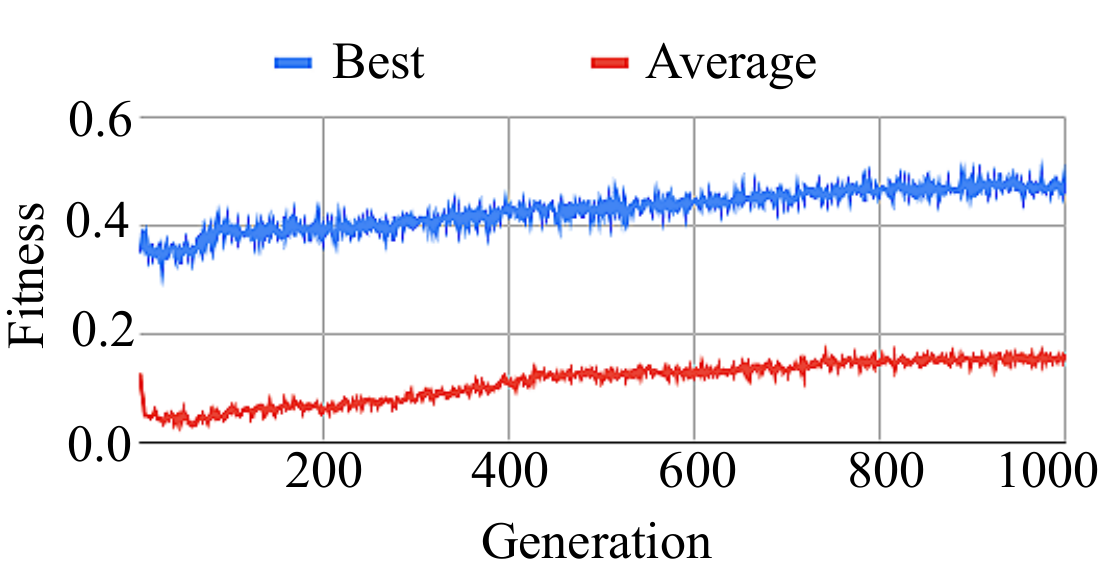}
        \caption{MC = 0.2}
        \label{fig:avg_fit_02}
    \end{subfigure}
    \begin{subfigure}[t]{.33\linewidth}
        \includegraphics[width=0.95\linewidth]{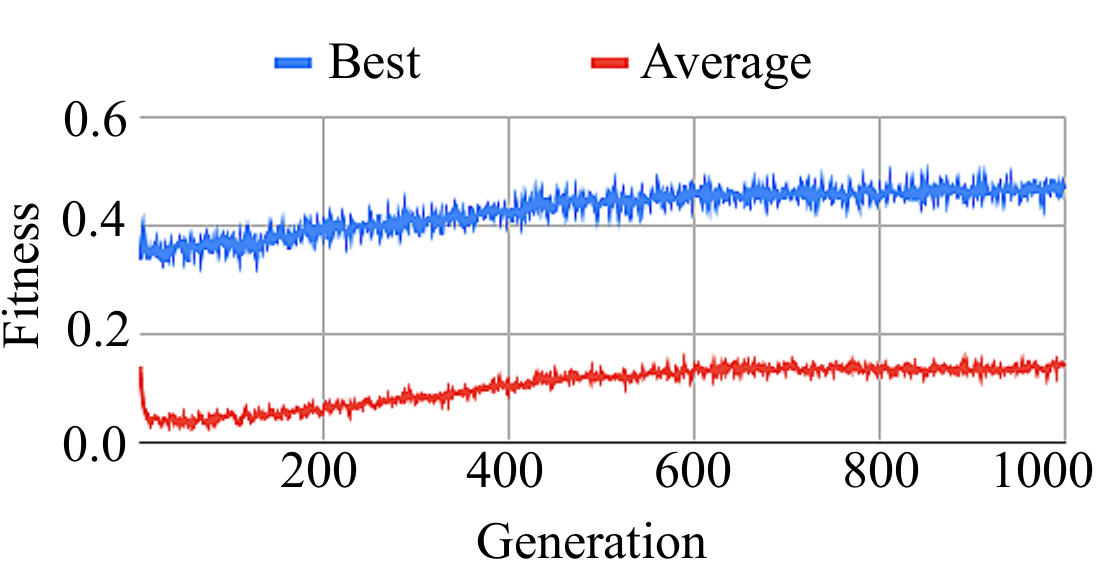}
        \caption{MC = 0.1}
        \label{fig:avg_fit_01}
    \end{subfigure}
    \begin{subfigure}[t]{.33\linewidth}
        \includegraphics[width=0.95\linewidth]{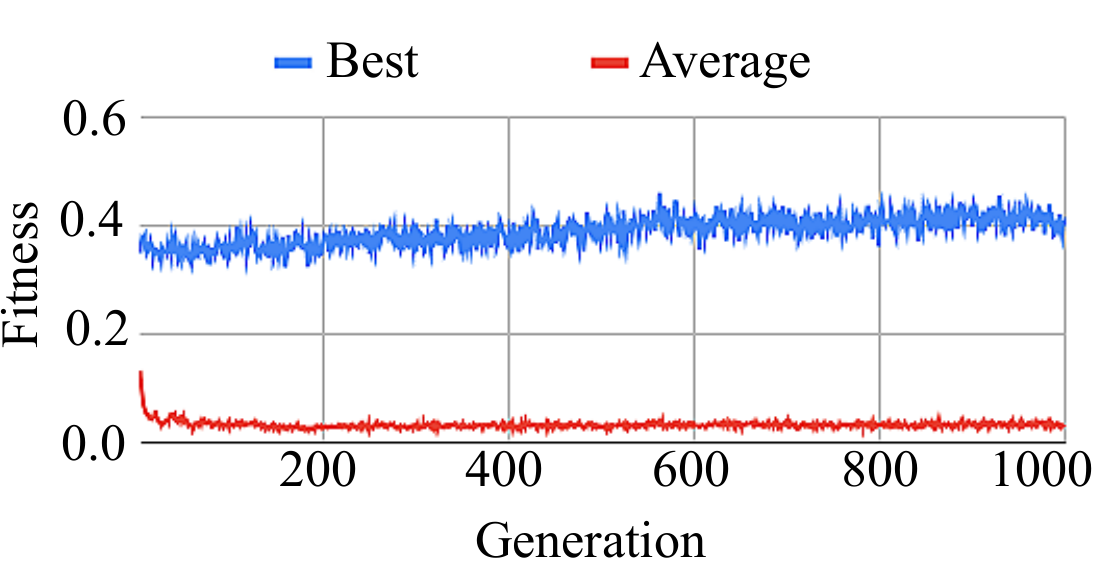}
        \caption{MC = 0.0}
        \label{fig:avg_fit_00}
    \end{subfigure}
    \caption{\textbf{Average (red) and best (blue) fitness over time.} Imposing \emph{no} minimal fitness requirement causes a stark decline in average fitness, highlighting the utility of minimal-criterion-based evolutionary algorithms when searching for collections of artifacts.}
    \label{fig:fitness}
\end{figure*}

\begin{figure*}
    \centering
    \begin{subfigure}[t]{.33\linewidth}
        \includegraphics[width=0.95\linewidth]{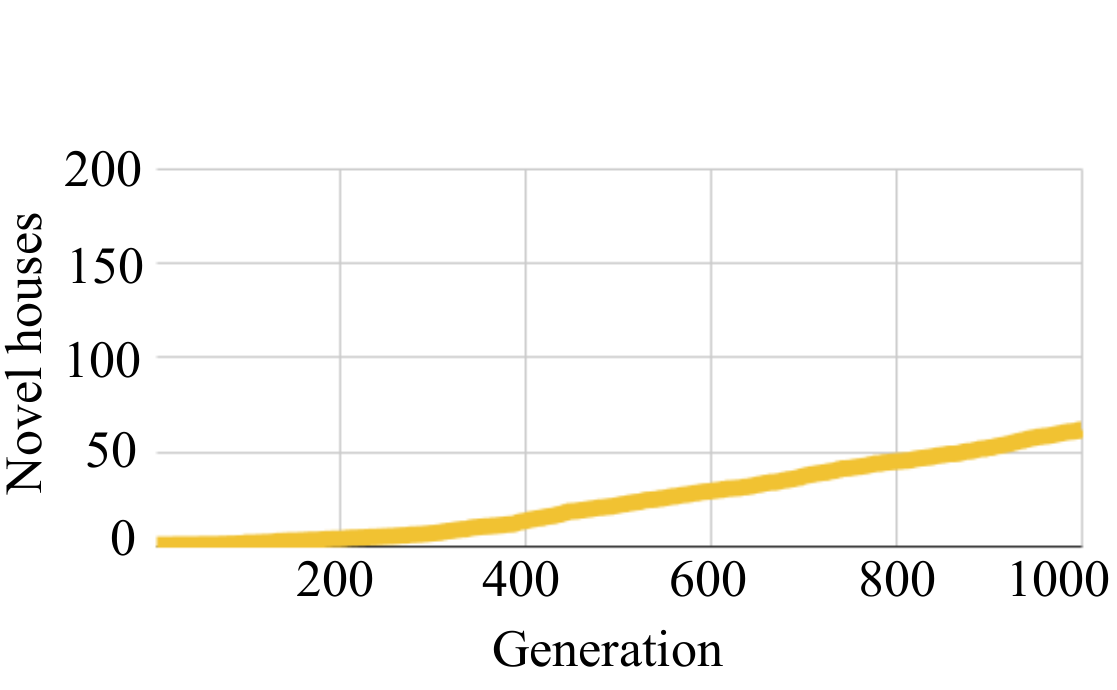}
        \caption{MC = 0.2}
        \label{fig:novel_02}
    \end{subfigure}
    \begin{subfigure}[t]{.33\linewidth}
        \includegraphics[width=0.95\linewidth]{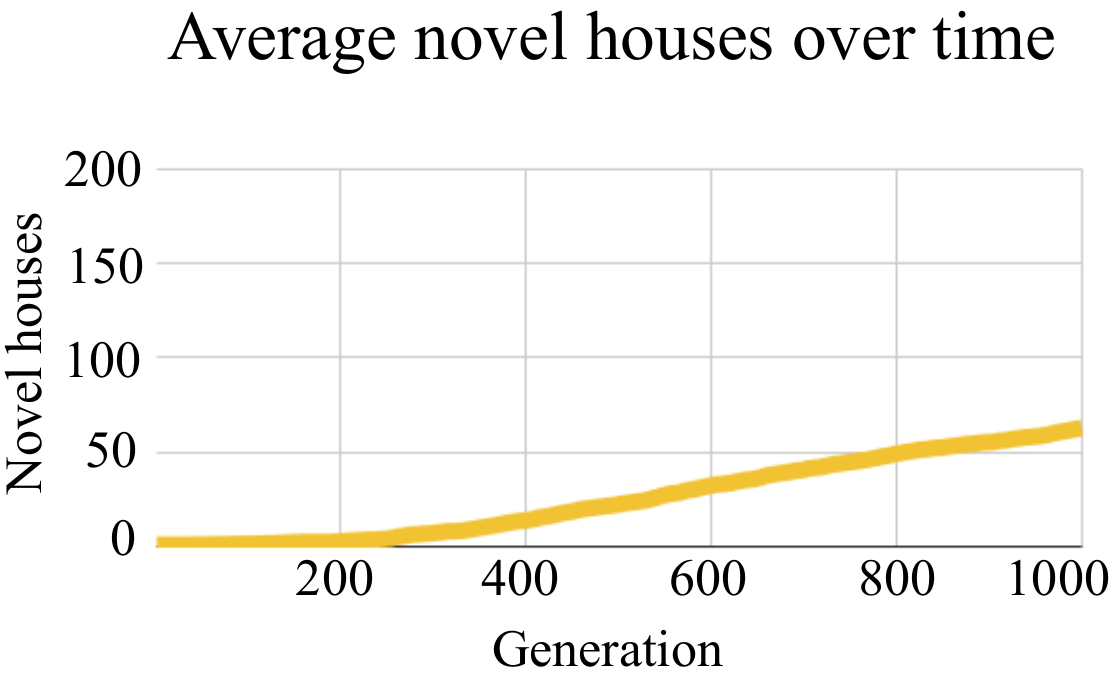}
        \caption{MC = 0.1}
        \label{fig:novel_01}
    \end{subfigure}
    \begin{subfigure}[t]{.33\linewidth}
        \includegraphics[width=0.95\linewidth]{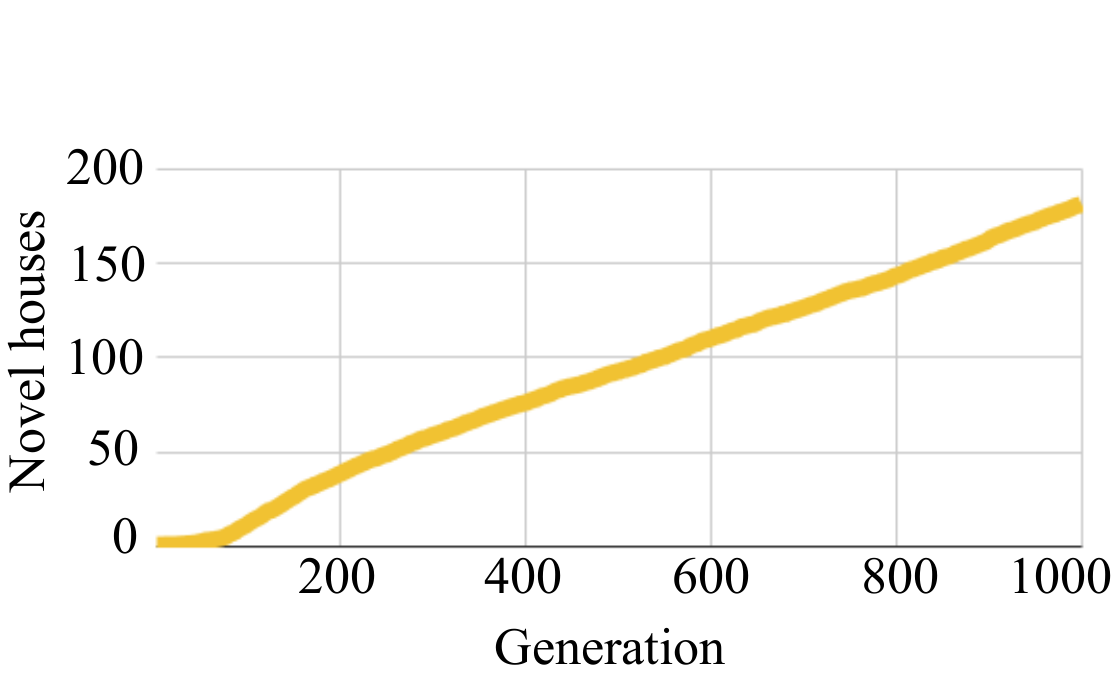}
        \caption{MC = 0.0}
        \label{fig:novel_00}
    \end{subfigure}
    \caption{\textbf{Average number of novel houses over time.} While pure novelty search finds more novel houses than versions with an imposed minimal criterion (MC), the effect of tested MC values is minimal.}
    \label{fig:novelty}
\end{figure*}

Table \ref{table:results} shows quantitative results for all runs. A successful algorithm applied to this domain would find a wide variety of houses that simple agents can live in. Figures \ref{fig:avg_fit_00} and \ref{fig:novelty} provide visual graphs of average house fitness over time and average number of novel houses found.

\begin{table}[!b]
  \centering
    \begin{tabular}{|p{2.3cm}||p{0.9cm}|p{1.0cm}|p{1.4cm}|p{0.7cm}|}
    \hline
    Result & NS & MCNS & MCNS-H & (1+1)\\
      \hline \hline
  Avg archive & 181.29 & 62.8 & 62 & N/A  \\
  Std dev & 25.94 & 32.44  & 25.31  & N/A  \\
  \hline
  Avg best fitness & 0.56 & 0.57 & 0.57 & 0.63 \\
  Std dev & 0.033  & 0.033  & 0.026  & .0163   \\
  \hline
  Best fitness & 0.63 & 0.63 & 0.63 & 0.66 \\
  
 \hline
    \end{tabular} \caption{\textbf{Experimental results.} The novelty search variants find a variety of houses with a slight quality sacrifice.}
    \label{table:results}
\end{table}


Figure \ref{fig:fitness} shows average fitness per generation in each experiment. With a minimum viability criterion implemented, the population's average fitness slowly increased each generation but began to level off around the 500th generation. At 1000 generations with 100,000 individuals, novelty searches with a minimum criteria value of 0.2 and 0.1 produced a population average fitness of 0.162 and 0.139 respectively. Compared to the novelty search with no minimum criteria value, the population average fitness dropped after the first generation and stayed low. This pure novelty search produced an average population fitness of 0.033. 

\begin{figure}[!h]
    \centering
    \includegraphics[width=0.48\textwidth]{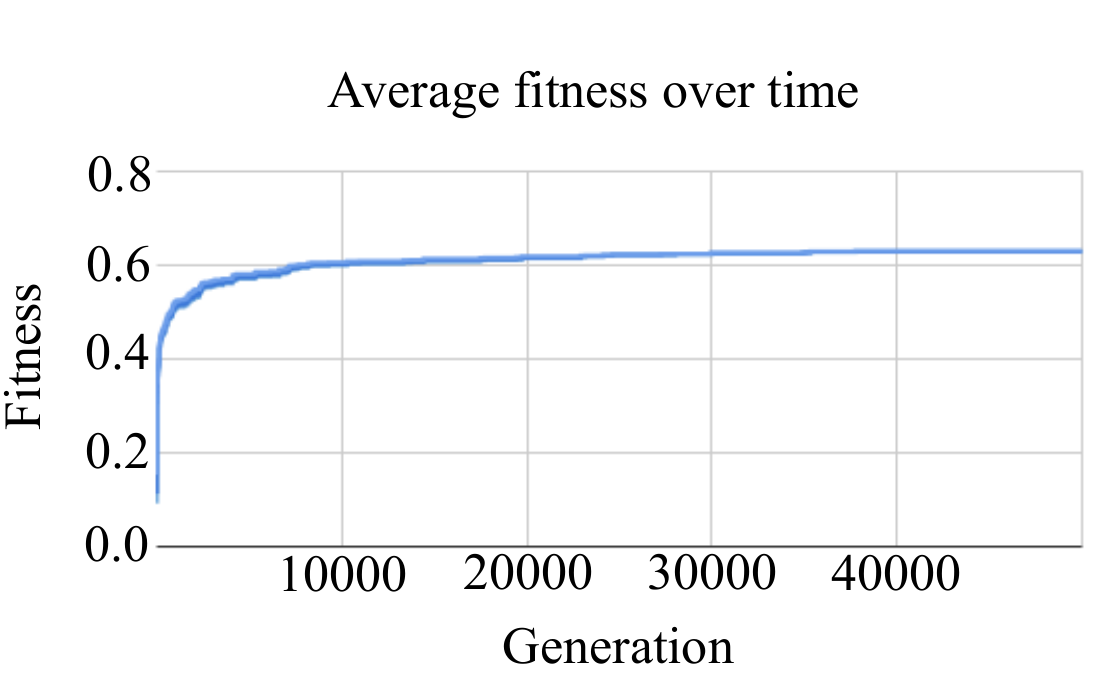}
    \caption{\textbf{(1+1) EA average fitness per generation}. This algorithm converges early in its run and only shows incremental improvement given more time. Note that the EA returns just \emph{one} interesting room instead of many.}
    \label{fig:1-1_fit}
\end{figure}

The blue lines in Figure \ref{fig:fitness} show the best fitness per generation for each MCNS experiment. The best fitness of each population gradually increased as they evolved and more novel houses were found and remained relatively unchanged for all three minimum criteria values. For the minimum criteria values of 0.2, 0.1, and 0.0 the best fitness from each population per generation was 0.568, 0.574, and 0.564 respectively. Average best fitness at end of run for minimum criteria values of 0.2, 0.1, and 0.0 was 0.568 ($\sigma=.0.026$), 0.574 ($\sigma=.0.033$), and 0.564 ($\sigma=.0033$) respectively.
For the (1+1) evolutionary algorithm, the average best fitness found for the final house generated was 0.63 ($\sigma=.0163$) with the maximum best fitness found being 0.667. Figure \ref{fig:1-1_fit} shows the average fitness per generation. 

While fitness information informs our understanding of whether or not the implemented algorithms sacrifice quality for the sake of diversity, the intended algorithmic challenge in this domain is for a single run of a search algorithm to find a diverse set of viable designs, not just a single best room. Figure \ref{fig:novelty} shows the number of novel houses found per generation of the three experiments. In the two experiments with algorithms incorporating a minimal fitness criterion, the population's average fitness slowly increased each generation. While the graphs could potentially imply that fitness is still increasing, a smaller set of preliminary experiments indicated that fitness actually begins to level off around the 500th generation. The average final number of novel houses found by the searches with minimum criteria values of 0.2 and 0.1 were 62 ($\sigma=25.31$) and 62.8 ($\sigma=32.44$) respectively, while the final number of novel houses for the search with a minimum criteria value of 0.0 was 181.28 ($\sigma=25.94$). No number of novel houses is reported for the (1+1) evolutionary algorithm because it does not maintain an archive.


\begin{figure}
\centering
  \includegraphics[width=0.157\textwidth]{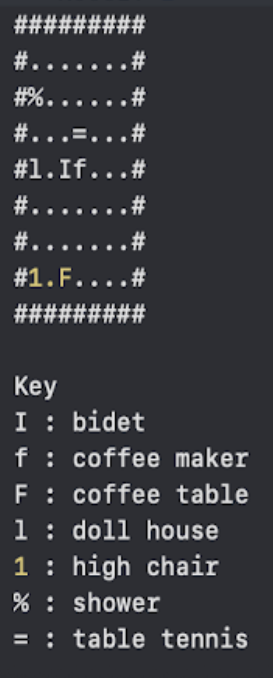}
  \includegraphics[width=0.2\textwidth]{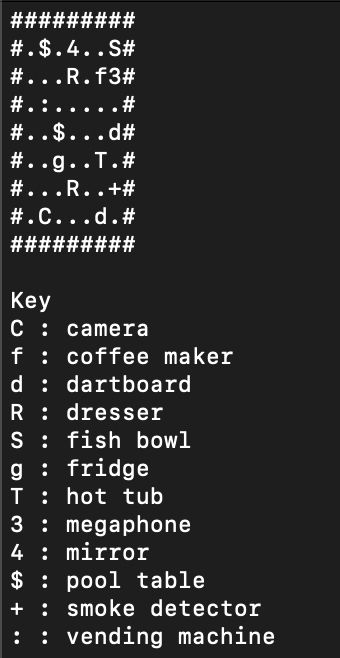}
  \caption{\textbf{Sample houses evolved by the (1+1)-EA and Novelty Search, respectively.} Houses are output as JSON that can then be rendered as ASCII. ``Optimal'' houses found by the greedy (1+1)-EA prefer a few specific maximally need-satisfying items, while ``novel'' houses include a wider variety of objects that still satisfy the Sim's needs. 
  }
  \label{fig:genroom1}
\end{figure}


\section{Discussion}





The purpose of the experiments in this paper was to investigate minimal criterion search in a domain that has a natural minimal criterion for success (avoiding death), which is a novel challenge tailored for divergent evolutionary search algorithms (which are becoming increasingly popular for PCG). This fitness criterion was then reinterpreted as a tunable parameter for exploring minimal-criterion-based divergent search. In the end, the fitness thresholds chosen did not have significant impacts on best fitness or number of houses found, but having \emph{some} minimal fitness criterion on top of pure novelty search did result in a higher average fitness. It is also interesting to note that the hillclimber tended to find the best fitness, suggesting that the design problem in its current form is not deceptive. However, finding a single house with high fitness, while interesting, frames the goal of evolution as solving a single-objective optimization problem. The fact still stands that a single run of the hillclimber is not nearly as generative as i.e. novelty-based methods, which discover entire archives of viable artifacts. None of the houses generated were able to reach a maximum fitness of 1.0 due to the fact that none of the placeable objects could max out any one need value for the agent. The Sims agent only interacted with an object enough to raise their diminished need above the set threshold value.

The system tended to replace the three starter objects in the house (a bed, toilet, and fridge) with more need-efficient objects included in the object list. This tendency demonstratseshow the algorithm was able to select which objects were the ``best'' objects for a house. For example, the ``coffee maker'' object replaced beds in 50\% of the final houses produced because the coffee maker renewed more of the Energy need than the bed object and also replenished Hunger. Similarly, in 25\% of the final houses, a ``bidet'' replaced the starting toilet object. While this object replenished the same amount of Bladder as the toilet, the bidet also replenished Fun and thus proved to be a more beneficial object in the house. Needs that were also not addressed by the starter objects, such as the Fun and Social needs, were also optimized through the hillclimber algorithm. In 70\% of the final houses, either a ``foosball table'' or ``table tennis'' object was placed because these objects best satisfied both the Social and Fun needs.

Finding the optimal objects in the object lists given to the mutators could help to identify any imbalances that might occur from including or excluding certain objects in the house. The case study presented in this paper adds to the ever-growing collection of domains where evolution, and specifically evolution-based procedural content generation, proves useful for automated analysis of game design. The notable recurrence of a few very specific furniture items in most high-performing houses would suggest, for instance, nerfing the coffee maker. Algorithms that are explicitly designed to find a diversity of high-performing artifacts are uniquely suited for this role.

In future work, a monetary constraint (like those used in the original Sims games when constructing houses) could be implemented to challenge the algorithm to evolve a house towards optimization given constraints on the object selection. This resource-constrained version of the design problem would provide an interesting challenge for quality diversity algorithms, and may in fact add deceptive elements to the domain. We would also like to experiment with the Sim agent behavior to see how overall house generation is affected. Such experiments could include changing the ranking of needs, allowing diagonal movement, replenishing a need to the maximum value, etc. In addition to adding evolution constraints and altering agent behavior, many other evolution-centric parameters (i.e. mutation probabilities, minimum k-nearest neighbor distance) could be played with to see how the ending novelty archive is affected.

\section{Conclusion}

This paper introduced a novel testbed for divergent search search and PCG based on the sandbox game The Sims. 
Various evolutionary algorithms were implemented as a means to explore house creation space and provide an initial foundation for future work in the sandbox game domain. These algorithms were able to generate hundreds of ``Sims-livable'' houses with a variety of furnished objects and additionally exposed potential imbalances in the game's design space. 


\section*{Acknowledgements}
This work was supported by the National Science Foundation (Award number 1717324 - “RI: Small: General Intelligence through Algorithm Invention and Selection.”). 

\bibliographystyle{aaai}
\bibliography{newrefs, References}

\end{document}